\documentclass{article}
\usepackage{spconf,amsmath,graphicx}
\usepackage{rotating}
\usepackage{multirow}

\title{Ensembling object detectors for image and video data analysis}
\name{Kateryna Chumachenko$^{\dagger}$, Jenni Raitoharju$^{\dagger \mathsection}$, Alexandros Iosifidis$^{\star}$, Moncef Gabbouj$^{\dagger}$ \thanks{This work was supported by Business Finland under the project 5G Vertical Integrated Industry for Massive Automation (5G-VIIMA).}}
\address{$^{\dagger}$Tampere University, Faculty of Information Technology and Communication Sciences, Finland  \\
$^{\mathsection}$ Finnish Environment Institute, Programme for Environmental Information, Finland \\
    $^{\star}$Aarhus University, Department of Electrical and Computer Engineering,  Denmark 
    }
\begin{document}
\ninept
\maketitle

\begin{abstract}
In this paper, we propose a method for ensembling the outputs of multiple object detectors for improving detection performance and precision of bounding boxes on image data. We further extend it to video data by proposing a two-stage tracking-based scheme for detection refinement. The proposed method can be used as a standalone approach for improving object detection performance, or as a part of a framework for faster bounding box annotation in unseen datasets, assuming that the objects of interest are those present in some common public datasets. 
\end{abstract}


\section{Introduction}
Driven by the broad availability of an extensive amount of datasets in different domains, object detection has become one of the most widely used tools within the field of computer vision in recent years, finding applications in various areas, such as video surveillance \cite{surv}, medical diagnostics \cite{cancer}, historical image analysis \cite{ww2}, and industrial applications \cite{apples}. 
Despite the huge progress made in the field of object detection in the recent years, not much attention has been paid to the generalization ability of the object detection methods: the developed algorithms assume that the training and test data come from the same distribution, and the test performance is reported on data from the same dataset used for training. 
For this reason, it is always preferable to train the methods on data collected directly from the domain of application, even if the classes of interest are present in some public datasets. Nevertheless, the models trained on public datasets are still often used in industrial applications. This is primarily due to the fact that obtaining training data from the domain of interest is generally both expensive and time-consuming, as it requires a significant amount of manual labour related to the annotation of the data specific to the application. 

\begin{figure}
\centering
    \includegraphics[width=0.48\textwidth]{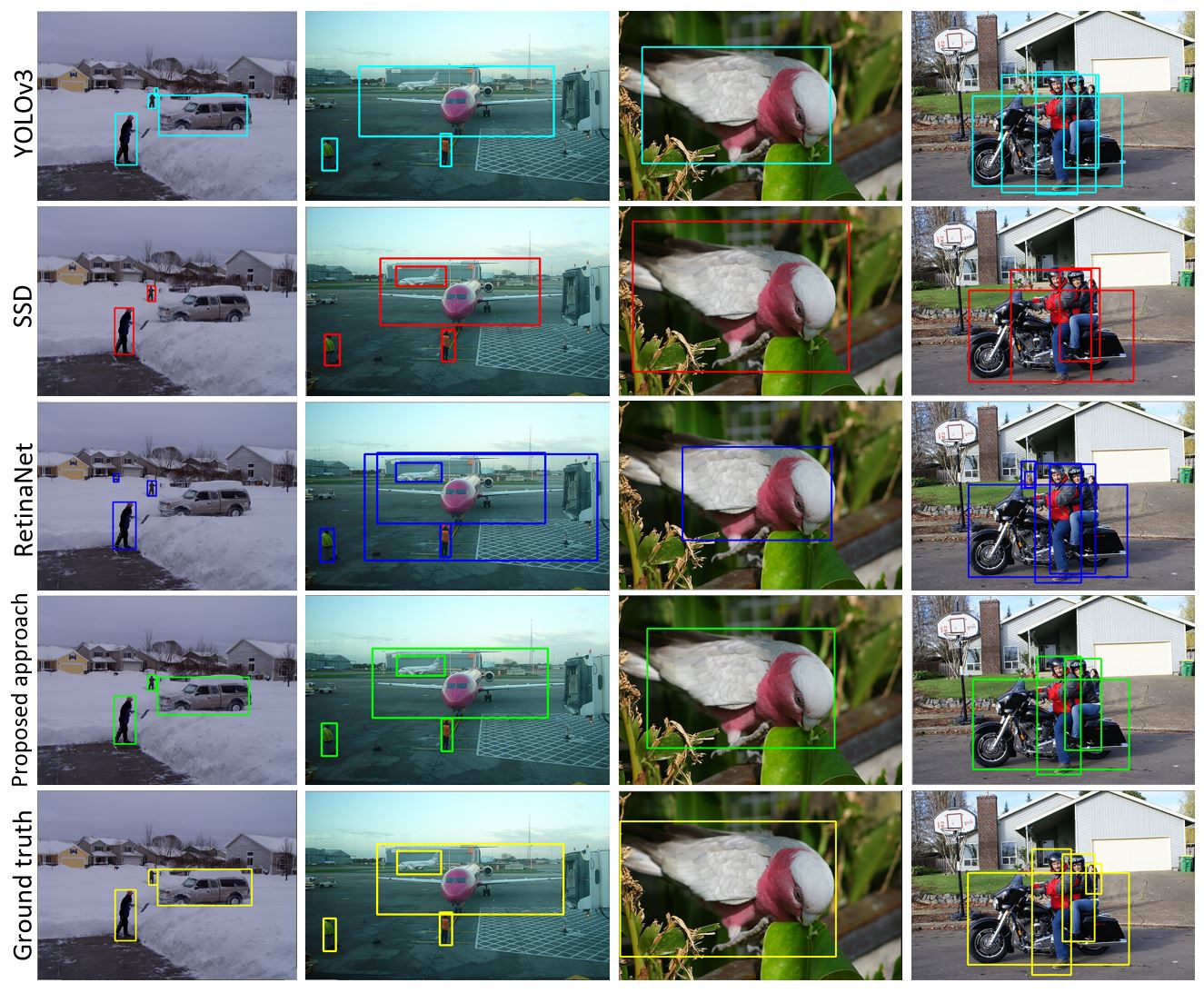}
    \caption{Examples of the detection results of base detectors, proposed method, and ground truth bounding boxes on PASCAL VOC dataset.}\label{fig:examples}
\end{figure}

Several methods have been proposed for speeding up the bounding box annotation process on images \cite{feifei,  bishwo,assistlearn}. Most of them still require large amounts of manual annotations, corrections, and retraining of the models. However, it is often the case that the classes of interest belong to the set of common classes present in public datasets, such as people, vehicles, and animals. Using widely-available object detection methods pre-trained on such public datasets has the potential to significantly reduce the amount of manual labour required for the annotation process. In this work, we aim to take a step towards utilizing such methods in a way that improves the object detection performance and reduces the annotation time. 

Despite the rapid development of object detection methods \cite{objdet,  ssd,yolo, retinanet}, only a very limited number of works have focused on creating ensembles of those for improved detection performance \cite{afa, npddpm,two2tango}. 
Moreover, to the best of our knowledge, none of the existing approaches aims at improving the precision of resulting bounding boxes, although precision can play a crucial role in a variety of applications, primarily including tracking and re-identification problems. 

In this paper, we aim to take a step towards improving object detection performance and precision in image and video data by proposing a method for ensembling multiple object detection methods. In addition, we propose a weighting scheme with regression-based weights learnt from a small number of images, as well as an extension for video data utilizing temporal information. 

\section{Related Work}\label{S:RelatedWork}
Significant progress has been made in the field of object detection in recent years \cite{objdet, ssd, yolo, retinanet}, dominated by deep learning-based methods. To improve the object detection performance, several approaches for combining information from multiple object detectors have been proposed recently.
In \cite{npddpm}, a cascade of two face detectors was proposed to reduce the number of false-positive detections. Several metrics for evaluating the diversity and correlation of detectors were used to select the best pair of detectors in \cite{two2tango}. In \cite{afa}, an SVM classifier was used to select the suitable detection method for speed considerations. A learning to rank based approach, intended to rank the more relevant detections higher, was proposed in \cite{detect2rank}. 
Notably, in all the above-mentioned approaches, the combination is aimed at the selection of the best detection. 
In our method, 
we first use contextual information of detectors to select the relevant detections and subsequently improve the precision of the final bounding box by taking a weighted average of the coordinates with the weights learned from a small subset of images of the target dataset.

The need for data annotation in the application domain motivates the emergence of various methods for speeding up the bounding box annotation process. A common approach to annotating large-scale datasets is using crowd-sourced annotations \cite{feifei}. Since this approach exposes the data to public, it cannot be used for applications where the data needs to be kept private. 
Another approach relies on self-training \cite{bishwo, assistlearn}. There, the general methodology is to first annotate a set of images manually and use them for the training of an object detector. The trained detector is subsequently used for producing the bounding boxes for the rest of the dataset. The obtained detections are manually refined, and the process continues until perfect annotation is achieved. 
Although minimizing to some extent the needed workload, the self-training approaches still require a high number of images to be annotated. In addition, a certain amount of time is needed for training the object detector. 

\section{Proposed Methods}\label{S:ProposedApproach}
The main motivation behind the proposed methods lies in the assumption that distinct object detectors achieve different levels of performance on different data, and extraction of useful information from each method has potential for improving the detection accuracy and precision of bounding boxes. The proposed approach can be used as a standalone methodology for improving object detection performance or as a part of a framework for creating bounding box annotations. 
We rely on three state-of-the-art object detectors: SSD \cite{ssd}, YOLOv3 \cite{yolo}, and  RetinaNet \cite{retinanet}. 
The choice of these detectors is based on their fast execution and good object detection performance reported in the literature. 
Here, one should note that the method can be equally well applied to any set of object detectors.

\subsection{Fusion by non-maximum suppression}
First, we consider the straightforward way of fusion by non-maximum suppression that will serve as a baseline for our work. Here, we treat all obtained detections as coming from one object detector and apply non-maximum suppression to suppress non-confident duplicates of detections identified to be the true positive ones. First, all bounding boxes detected in the image are sorted according to their confidence scores. The most confident bounding box is selected as the true detection. Then, the Intersection over Union $ IoU = \frac{Area\:\:of\:\:overlap}{Area\:\:of\:\:union}$ between the true detection's bounding box and every other detection's bounding box is calculated. The bounding boxes with higher IoU (above a certain threshold $\theta$, equal to $0.5$ in our work) are identified as belonging to the same object as the true bounding box and removed from the list of detections. The process continues from the next most confident bounding box out of the remaining ones and is performed for each class separately.

Although providing a reasonable level of performance improvement, this approach suffers from several limitations. Firstly, the final bounding boxes are simply bounding boxes detected by one of the base detectors, limiting the potential of fusing the knowledge of multiple methods. Secondly, the use of confidence score of the detector as a metric for deciding the quality of bounding boxes is questionable. The common interpretation of the confidence score is the probability of the bounding box to be the true positive one. However, such interpretation does not have a strong theoretical basis, as confidence scores of different object detection methods, being learned parameters, can reflect significantly different true-positive rates. In other words, the confidence score scales of different object detection approaches are not calibrated, and some methods tend to produce high-confidence detections that are false-positives, while others output true positive detections with low confidence scores. 

\subsection{Proposed method for ensembling bounding boxes}
Taking a step towards more meaningful use of the knowledge provided by several object detectors, we propose the following approach that adjusts the final selection of bounding boxes based on how likely they are to correspond to true detections. First, the previously described IoU-based merging is employed to identify the bounding boxes belonging to the same object. However, instead of discarding all the bounding boxes corresponding to the same object as the most confident bounding box, we keep track of the source detectors for each bounding box. From the set of bounding boxes obtained from each detector, we select only the one having the highest IoU with the most confident bounding box. As a result, we obtain a set of detections, each described by 1-3 bounding boxes. At this point, we can observe that objects detected by only one of the detectors are generally false-positives, so out of the obtained set of detections, we discard the ones described by only one bounding box.
In order to obtain the final detections, the selected bounding boxes for each detection need to be combined. To exploit the knowledge present in each of these bounding boxes, we propose to fuse them as a weighted combination with learned weights.

\subsubsection{Improving precision of bounding boxes}
In order to improve the precision of the bounding box coordinates, we further extend the proposed approach by taking a weighted average value of the coordinates of each of the vertices of the bounding boxes identified to belong to the same object. We weight each coordinate by the corresponding confidence score of the detection and normalize the resulting value by the sum of the confidence scores so that more confident detectors put more weight to the final output. Direct use of confidence scores would suffer from the limitations caused by the scores of different detectors not being calibrated. To address this issue, we propose a scheme for re-weighting the output of each detector that results in better calibration of confidence scores, and, therefore, in a more meaningful combination of the bounding boxes. Assuming that we have obtained a scalar weight $w^j$ for the $j^{th}$ detector, the new coordinates and confidence scores are calculated as
\begin{eqnarray}
\hat{c_i} =\frac{\sum_{j=1}^D{s^j_iw^jc^j_i}}{\sum_{i=1}^{D}{s^j_iw^j}} \:\:\:\:\:\:\:\:\: \textrm{and} \:\:\:\:\:\:\:\:\:
\hat{s_i} = \frac{\sum_{j=1}^D{w^js^j_i}}{\sum_{j=1}^D{w^j}},
\end{eqnarray}
where $\hat{c_i}$ and $\hat{s_i}$, are the refined coordinate and updated confidence score of $i^{th}$ detection, $D$ is the number of detectors,  $w^j$ is the weight of $j^{th}$ detector, and $s_i^j$ and $c_i^j$ are the score and coordinate of $i^{th}$ detection of $j^{th}$ detector.

Appropriate weights $w^j$ for each detector cannot be determined empirically. Therefore, we formulate a regression problem to learn these weights from a low number of manually annotated images. Let us denote by $\mathbf{b}_i^{j} = [x^{j}_{i1}, y^{j}_{i1}, x^{j}_{i2}, y^{j}_{i2}]$ a 
vector of coordinates representing the $i^{th}$ bounding box of detector $j$, where $x^{j}_{i1}$, $y^{j}_{i1}$, $x^{j}_{i2}$ and $y^{j}_{i2}$ are the coordinates of two of the bounding box corners. Let the confidence score corresponding to this bounding box be $s^{j}_i$ and $\mathbf{g}_i = [g_{x1}, g_{y1}, g_{x2},  g_{y2}]$ be the corresponding groundtruth vector of coordinates. Assuming that we are operating with $D$ distinct object detection methods, each $i^{th}$ detection $\mathbf{X}_{i}$ can be represented by $D\times4$ matrix $\mathbf{X}_{i} = [\mathbf{b}_i^{1}s_i^{1}; \mathbf{b}_i^{2}s_i^{2}; ...; \mathbf{b}_i^{D}s_i^{D}]$ of confidence score-scaled bounding box coordinates obtained from $D$ detectors. Therefore, our goal is to find a $D\times1$ dimensional set of weights $\mathbf{w} = [w^{1}, w^{2},..., w^{D}]^T$  that would satisfy the following criterion for each detection $i$:
\begin{equation}
\centering
\begin{gathered}
\mathbf{w}^{T}\mathbf{X}_{i} = \mathbf{g}_{i}  \Rightarrow
\begin{array}{c}
[w^{1}, w^{2},..., w^{D}]
\end{array} \times
\left[ \begin{array}{cc}
\mathbf{b}_i^{1} s_i^{1}\\
    \mathbf{b}_i^{2} s_i^{2}\\
    ...\\
    \mathbf{b}_i^{N} s_i^{D}\\
\end{array} \right] = 
\begin{array}{cc}
\mathbf{g}_i,
\end{array}
\end{gathered}
\notag
\end{equation}
\begin{equation}
    \begin{gathered}
\begin{array}{c}
[w^{1}, w^{2},..., w^{D}]
\end{array} \times
\left[ \begin{array}{cccc}
x^{1}_{i1}s_i^{1} & y^{1}_{i1}s_i^{1}  & x^{1}_{i2}s_i^{1} & y^{1}_{i2}s_i^{1} \\
  x^{2}_{i1}s_i^{2}& y^{2}_{i1}s_i^{2}& x^{2}_{i2}s_i^{2}&  y^{2}_{i2}s_i^{2} \\
    ... & ... & ... & ...\\
    x^{D}_{i1}s_i^{D}& y^{D}_{i1}s_i^{D}& x^{D}_{i2}s_i^{D}& y^{D}_{i2}s_i^{D}\\
\end{array} \right] 
= \\
= \left[ \begin{array}{cccc}
g_{x1} & g_{y1} & g_{x2} & g_{y2}
\end{array} \right].
\end{gathered}
\end{equation}

To obtain the solution to the problem, we optimize it iteratively by means of Stochastic Gradient Descent optimized over the Mean Squared Error (MSE), defined as 
\begin{equation}
\mathcal{L}_{MSE} = \frac{1}{4n}\sum_{i=1}^n\sum_{d=1}^4(\mathbf{g}_{i,d} - \hat{\mathbf{g}}_{i,d})^2,
\end{equation} where $n$ is the number of training samples, $\mathbf{g}_{i,d}$ is the $d^{th}$ coordinate of the ground truth bounding box of $i^{th}$ sample, and $\hat{\mathbf{g}}_{i,d}$ is the corresponding predicted coordinate. \subsubsection{Creating detection-ground truth pairs}

The described process requires the creation of a training set of detection-ground truth pairs $\{\mathbf{X}_i; \mathbf{g}_i\}$. For this purpose, we first manually annotate a low number of images with the ground truth bounding boxes (100 in our experiments), as annotation of such a small number of images is not time-consuming, and apply the base object detection methods on these images. 
To reduce the amount of manual labour required for manually assigning each bounding box to the corresponding ground truth box, for forming the $\{\mathbf{X}_i; \mathbf{g}_i\}$ pairs we follow the following process: for each ground truth bounding box $\mathbf{g}_i$ we find all the overlapping bounding boxes of the corresponding class obtained by different detectors and select the ones having the IoU higher than a certain threshold $\theta$. From this set, for each of the detectors we select the bounding box having the highest IoU with the ground truth box, and the rest of the bounding boxes are discarded. The obtained bounding boxes are then used to create the matrix $\mathbf{X}_i$ corresponding to the ground truth coordinates~$\mathbf{g}_i$. In case one of the detectors did not produce a detection that would correspond to the groundtruth, we simply set the corresponding bounding box in $\mathbf{X}_i$ to zeros, i.e., $\mathbf{b}_i^j = [0,0,0,0]$. 
We discard  redundant duplicate detections and false-positive detections that did not match with any of the groundtruth bounding boxes, since the re-weighting of bounding boxes cannot fix the presence of incorrect detections as such. 

\begin{table*}[h!]
\caption{MAP results on PASCAL VOC dataset}
\footnotesize
\setlength{\tabcolsep}{1.5pt}
\renewcommand\arraystretch{0.95}

\begin{tabular}{|l|l|c|c|c|c|c|c|c|c|c|c|c|c|c|c|c|c|c|c|c|c|c|}
\hline
                         & \textbf{}                                                                                 & \textbf{plane} & \textbf{\textbf{bicycle}} & \textbf{\textbf{bird}} & \textbf{\textbf{boat}} & \textbf{\textbf{bottle}} & \textbf{bus}   & \textbf{\textbf{car}} & \textbf{\textbf{cat}} & \textbf{\textbf{chair}} & \textbf{\textbf{cow}} & \textbf{table} & \textbf{\textbf{dog}} & \textbf{\textbf{horse}} & \textbf{\textbf{bike}} & \textbf{\textbf{person}} & \textbf{plant} & \textbf{sheep} & \textbf{sofa}  & \textbf{train} & \textbf{monit.} & \textbf{TOTAL} \\ \hline
\multirow{5}{*}{\rotatebox[origin=c]{90}{\textbf{IoU 0.5}}}  & {\textbf{Ret.Net} }                                                    & 86.28                               & 81.55                                      & 75.01                                   & 54.16                                   & 63.04                                     & 80.08                           & 82.05                                  & 88.71                                  & 57.13                                    & 79.61                                  & 57.47                           & 81.65                                  & 82.7                                     & 84.02                                   & 82.59                                     & 48.49                           & 72.17                           & 65.41                           & 84.25                           & 73.97                            & 74.02                           \\
                         & \textbf{\textbf{ {\textbf{SSD}}}}    & 82.8                                & 80.14                                      & 72.09                                   & 52.55                                   & 51.78                                     & 81.56                           & 77.75                                  & 88.49                                  & 56.24                                    & 76.64                                  & 61.32                           & 81.47                                  & 83.84                                    & 82.35                                   & 79.39                                     & 41.62                           & 70.75                           & 65.82                           & 87.85                           & 71.13                            & 72.28                           \\
                         & \textbf{\textbf{ {\textbf{Yolov3}}}} & 88.33                               & 83.88                                      & 69.2                                    & 55.56                                   & 65.12                                     & 86.34                           & 80.67                                  & 86.79                                  & 66.76                                    & 70.49                                  & 66.62                           & 82.51                                  & 87.32                                    & 84.19                                   & 83.24                                     & 50.02                           & 67.9                            & 71.91                           & 86.59                           & 74.96                            & 75.42                           \\
                         & \textbf{\textbf{ {\textbf{NMS}}}}    & 90.92                               & 86.57                                      & \textbf{78.94}         & 63.79                                   & \textbf{69.38}           & \textbf{87.11} & 84.88                                  & \textbf{91.88}        & \textbf{68.83}          & 82.43                                  & 66.72                           & 86.84                                  & 89.14                                    & 87.05                                   & 86.16                                     & \textbf{54.37} & \textbf{79.00} & \textbf{74.71} & 90.1                            & 78.65                            & 79.87                           \\
                         & \textbf{\textbf{ {\textbf{Our}}}}    & \textbf{90.97}     & \textbf{87.08}            & 78.59                                   & \textbf{65.54}         & 68.45                                     & 87.08                           & \textbf{84.92}        & 91.85                                  & 68.47                                    & \textbf{84.16}        & \textbf{67.14} & \textbf{87.48}        & \textbf{89.94}          & \textbf{87.46}         & \textbf{86.57}           & 53.38                           & 78.69                           & 74.33                           & \textbf{90.38} & \textbf{79.06}  & \textbf{80.08} \\ \hline
\multirow{5}{*}{\rotatebox[origin=c]{90}{\textbf{IoU 0.75}}} & \textbf{{\textbf{Ret.Net} }}                          & 60.66                               & 50.38                                      & 48.36                                   & 24.06                                   & 31.87                                     & 71.06                           & 59.31                                  & 71.03                                  & 33.62                                    & 60.35                                  & 33.86                           & 59.62                                  & 58.45                                    & 59.14                                   & 46.75                                     & 19.08                           & 44.44                           & 50.53                           & 67.11                           & 50.95                            & 50.05                           \\
                         & \textbf{ {\textbf{SSD}}}                              & 56.11                               & 46.80                                      & 42.79                                   & 22.20                                   & 23.82                                     & 70.51                           & 55.95                                  & 65.67                                  & 28.41                                    & 56.50                                  & 34.46                           & 56.89                                  & 54.79                                    & 53.67                                   & 41.92                                     & 12.14                           & 46.89                           & 45.08                           & 65.33                           & 47.66                            & 46.38                           \\
                         & \textbf{ {\textbf{Yolov3}}}                           & 56.67                               & 51.29                                      & 43.31                                   & 22.29                                   & 38.19                                     & 76.59                           & 56.87                                  & 65.98                                  & 38.36                                    & 52.12                                  & 38.87                           & 60.42                                  & 60.53                                    & 55.10                                   & 51.57                                     & 18.93                           & 41.49                           & 52.59                           & 63.37                           & 48.36                            & 49.64                           \\
                         & \textbf{ {\textbf{NMS}}}                              & 62.72                               & 53.85                                      & 51.46                                   & 27.74                                   & 39.38                                     & 77.11                           & 60.85                                  & 71.17                                  & 39.42                                    & 64.30                                  & 39.15                           & 63.89                                  & 62.82                                    & 58.78                                   & 53.26                                     & 19.71                           & 52.93                           & 55.66                           & 69.13                           & 51.52                            & 53.75                           \\
                         & \textbf{ {\textbf{Our}}}                              & \textbf{67.43}     & \textbf{59.32}            & \textbf{54.43}         & \textbf{31.65}         & \textbf{42.69}           & \textbf{78.32} & \textbf{64.41}        & \textbf{77.37}        & \textbf{43.70}          & \textbf{66.41}        & \textbf{43.55} & \textbf{69.74}        & \textbf{66.24}          & \textbf{64.43}         & \textbf{56.60}           & \textbf{22.84} & \textbf{55.21} & \textbf{60.05} & \textbf{72.39} & \textbf{60.04}  & \textbf{57.84} \\ \hline
\multirow{5}{*}{\rotatebox[origin=c]{90}{\textbf{IoU 0.85}}} & {\textbf{Ret.Net} }                                                    & 36.75                               & 28.57                                      & 24.17                                   & 7.56                                    & 10.12                                     & 56.36                           & 34.60                                  & 48.50                                  & 14.55                                    & 33.28                                  & 17.4                            & 40.01                                  & 36.60                                    & 31.75                                   & 20.83                                     & 5.64                            & 21.65                           & 32.27                           & 45.52                           & 21.81                            & 28.40                           \\
                         & \textbf{ {\textbf{SSD}}}                              & 27.31                               & 16.25                                      & 16.70                                   & 5.11                                    & 7.71                                      & 55.51                           & 27.50                                  & 33.09                                  & 9.47                                     & 29.89                                  & 11.99                           & 28.30                                  & 29.11                                    & 19.72                                   & 14.33                                     & 1.95                            & 23.10                           & 21.57                           & 32.45                           & 15.97                            & 21.35                           \\
                         & \textbf{ {\textbf{Yolov3}}}                           & 18.68                               & 19.32                                      & 15.90                                   & 5.04                                    & 11.11                                     & 53.56                           & 27.69                                  & 35.84                                  & 13.42                                    & 22.75                                  & 12.28                           & 28.06                                  & 30.99                                    & 19.52                                   & 20.91                                     & 4.17                            & 15.14                           & 24.72                           & 32.64                           & 15.53                            & 21.36                           \\
                         & \textbf{ {\textbf{NMS}}}                              & 34.10                               & 22.07                                      & 24.30                                   & 5.97                                    & 11.79                                     & 56.55                           & 31.00                                  & 43.30                                  & 13.73                                    & 34.55                                  & 13.11                           & 35.02                                  & 36.99                                    & 25.04                                   & 21.60                                     & 4.14                            & 25.46                           & 26.92                           & 43.73                           & 18.93                            & 26.42                           \\
                         & \textbf{ {\textbf{Our}}}                              & \textbf{37.65}     & \textbf{30.51}            & \textbf{27.78}         & \textbf{10.53}         & \textbf{15.13}           & \textbf{63.76} & \textbf{40.11}        & \textbf{50.58}        & \textbf{19.13}          & \textbf{37.47}        & \textbf{21.04} & \textbf{44.65}        & \textbf{43.4}           & \textbf{35.38}         & \textbf{27.08}           & \textbf{6.05}  & \textbf{29.91} & \textbf{37.05} & \textbf{46.58} & \textbf{23.1}   & \textbf{32.35}                           \\ \hline
\end{tabular}
\end{table*}
\begin{table}[htp!]
\centering
\caption{MAP@0.5 results on video datasets.}
\setlength{\tabcolsep}{2pt}
\renewcommand\arraystretch{0.9}
\begin{tabular}{|l|cccccc|c|}
\hline
       & \small{\textbf{RetinaNet}} &\small{ \textbf{SSD}}    & \small{\textbf{YOLOv3}} &\small{ \textbf{NMS}}   & \small{\textbf{Our}}   & \small{\textbf{Our-tr.}}    \\
       \hline
\footnotesize{\textbf{EPFL Camp}} & 67.89     & 65.08  & 68.12  & 67.38 & 70.47  & \textbf{70.70} \\ 
\footnotesize{\textbf{EPFL Lab-6}}    & 91.24     & 90.48  & 88.80   & 92.12 & 92.21 & \textbf{92.53} \\
\footnotesize{\textbf{Campus Aud}}   & 79.97     & 76.39 & 71.34  & 80.1  & 81.14 & \textbf{82.36} \\
\hline
\end{tabular}
\end{table}

\subsection{Exploiting temporal information in videos}
An additional step can be taken to further improve the performance on video data, by taking advantage of temporal information. We can safely assume that objects in a video sequence do not appear randomly at different frames, but follow certain trajectories of considerable time length. Therefore, we can enrich the set of detections obtained using the proposed ensembling approach with the ones that were likely to be missed and reduce the number of false-positive ones by following a two-stage tracking-based approach. 

In the first stage, to obtain a set of detections that were likely missed by the object detectors, we apply a set of correlational trackers \cite{corr}. At the first frame, an object tracker is initialized from each of the detections of that frame and they are tracked throughout the video. At each subsequent frame, the tracked bounding boxes are matched with the detections of the frame following the IoU-based matching process. A successful match is defined by an IoU exceeding 0.5, or 0.4 in the case that the tracklet was initialized in the past 3 frames, due to the assumption that the shape of the object changes significantly when entering the field of view, leading to higher differences between the bounding boxes detected by the detectors and the one tracked by the tracker. 
For the detections not matched with any of the tracklets, a new tracker is initialized, and objects that were tracked successfully but for which no detections were found are continued to be tracked unless one of the following holds:
\begin{enumerate}
    \item the tracklet was only matched with detections for a small number of frames (in our experiments we set this number to 5) and then missed for more than 5 frames,
    \item the tracklet was not matched with any of the detections for more than 50 frames.
\end{enumerate}
The first rule is intended for discarding tracklets that are likely to be initialized from false-positive detections, and the second rule is for discarding tracklets corresponding to objects that are no more present in the scene, but for which tracking continued. When an object that was tracked, but not matched with detections for a certain number of frames is rematched with a detection again, the bounding boxes predicted by the tracker at the frames where no detection happened are added to the set of detections. 

At the second stage, a multi-object tracker \cite{deepsort} is applied to reduce the number of false-positive detections: using the set of detections enriched with the ones obtained from the tracker at the first stage, we identify the sequences of detections belonging to the same object. The resulting sequences that consist only of a small number of consecutive frames (less than 5 in our work) and thus the ones that are most likely corresponding to false-positive detections are discarded. Note that the choice of tracking methods here is dictated by their fast speed and any other tracking methods can be employed as well. 

\section{Experimental setup and results}

To assess the applicability of the proposed approaches to real-world problems and evaluate the ability of the base detectors to generalize to previously unseen data, we evaluate the algorithms on different datasets than they were trained on. We select three state-of-the-art object detectors as our base methods: SSD \cite{ssd} trained on images resized to $512\times512$, YOLOv3 \cite{yolo} on $416 \times 416$ images, and RetinaNet \cite{retinanet} on images rescaled such that the smaller side is equal to 800 pixels. All models are trained on MS-COCO dataset \cite{coco}. To evaluate the ability of the methods to generalize to previously unseen data, we report the results on the intersection of classes with previously unseen PASCAL VOC dataset (training + test set) \cite{pascal} and three video datasets: EPFL Campus-7 dataset, EPFL Lab-6 dataset \cite{epfl}, and Campus Auditorium dataset \cite{campusaud, campusaud2}. In all video datasets, only the people class is considered, since the groundtruth data is available only for this class. We use the Mean Average Precision (MAP) as a metric for evaluation of the detection performance as defined in PASCAL VOC 2012 challenge \cite{mapimpl, mapgit}. Out of all the obtained detections, we discard the ones with a confidence score below 0.05 and report the MAP at the default IoU of 0.5 for each class separately as well as the total MAP. Besides, to evaluate the effect of the proposed approach on the precision of the bounding boxes, we report MAP at higher IoUs of 0.75 and 0.85 for each class in the image dataset. In our experiments, we perform 5-fold cross-validation. In the image dataset, 100 samples are used for training the score re-weighting model in each of the folds, with the remaining 9863 images used for testing. In the video dataset, we split the videos into 5 continuous segments, in each of which the last 100 images are used for training with the rest of the frame sequence used for testing. This is done because the proposed tracking-based approach requires an uninterrupted video sequence. The same test sets are used for reporting the results of separate object detectors and the mean MAP value across 5 folds is reported. For learning the weights, the bounding box coordinates $x$ and $y$ are scaled by the image width and height, respectively. From the resulting pairs obtained from 100 annotated images, 30\% are taken for validation, and the regression model is trained on the remaining 70\% with a learning rate of $10^{-5}$ starting from zero-initialized weights. The MSE is calculated on the validation set at each iteration and training proceeds until MSE stops improving for a number of iterations, after which the weights resulting in the best performance are selected. We report separately the results of each object detection method, the proposed approach, and the proposed approach refined by the tracking-based refinement scheme in the video datasets. We also report the results obtained by applying solely the non-maximum suppression to the detectors' output to showcase that the improvement of the detection performance is caused by the re-weighting scheme to a large extent. 

The results on the object detection methods as well as the proposed approaches on image and video datasets are presented in Tables 1 and 2, respectively, where the best MAP is highlighted in bold.
On the PASCAL VOC dataset, the best overall performance among the base detectors is achieved by YOLOv3 at MAP@0.5, and by RetinaNet at MAP@0.75. This allows us to conclude that YOLOv3 is able to detect the presence of objects of interest in general better, but RetinaNet produces the bounding boxes that match with the ground truth boxes more closely. This observation also reinforces the motivation of our proposed approach - by combining outputs of several object detectors we can combine the fair detection ability of less precise detectors such as YOLOv3, but compensate the precision of bounding boxes by more precise detectors such as RetinaNet.
We observe that the proposed approach outperforms the base detectors with all the IoU thresholds. At MAP@0.5 the improvements range from $0.52\%$ for the dining table class up to $9.98\%$ for the boat class. At this IoU threshold our proposed approach is performing on par with non-maximum suppression, with overall improvement of $0.21\%$. However, the performance differences are increased with the increase in IoU threshold, - we achieve $4.09\%$ and $5.93\%$ better performance on than non-maximum suppression on MAP@0.75 and MAP@0.85, respectively. Besides, we outperform all base detection methods on all IoU thresholds, leading to $4.66\%$, $7.79\%$, and $3.95\%$ MAP improvements on IoU of 0.5, 0.75, and 0.85, respectively. Some example results on PASCAL VOC dataset can be seen in Figure 1. Note that for clarity only the detections with confidence score of at least 0.4 are shown. 
In the video dataset, 
the proposed regression-based approach achieves a significant improvement over all of the detectors on all three datasets, and applying the refinement process based on tracking pushes this improvement further, leading to an overall improvement of $2.58\%$, $1.29\%$, and $2.39\%$ on EPFL Campus-7, EPFL Lab-6, and Campus Auditorium datasets, respectively.

\section{Conclusions}
We proposed a method for ensembling multiple object detectors that re-weights the confidence scores and bounding box coordinates, as well as exploits contextual information. The method resulted in better MAP scores compared to base detectors, where the gap is especially large on higher IoU thresholds. The extention of the proposed method for video data utilizing temporal information pushes the improvement in performance even further. The proposed methods can, therefore, be utilized directly for obtaining improved object detection results or as a part of a framework for creating annotations on new datasets.
\bibliographystyle{IEEEbib}
\bibliography{refs}

\end{document}